\makeatletter
\newcommand{\dontusepackage}[2][]{%
  \@namedef{ver@#2.sty}{9999/12/31}%
  \@namedef{opt@#2.sty}{#1}}
\makeatother
\dontusepackage{subfigure}

\documentclass[]{article}

\usepackage{lmodern}
\usepackage{amssymb,amsmath}
\usepackage{ifxetex,ifluatex}
\usepackage[usenames,dvipsnames]{color}
\usepackage{fixltx2e} 
\ifnum 0\ifxetex 1\fi\ifluatex 1\fi=0 
  \usepackage[T1]{fontenc}
  \usepackage[utf8]{inputenc}
\else 
  \ifxetex
    \usepackage{mathspec}
    \usepackage{xltxtra,xunicode}
  \else
    \usepackage{fontspec}
  \fi
  \defaultfontfeatures{Mapping=tex-text,Scale=MatchLowercase}
  
\fi
\IfFileExists{upquote.sty}{\usepackage{upquote}}{}
\IfFileExists{microtype.sty}{%
\usepackage{microtype}
\UseMicrotypeSet[protrusion]{basicmath} 
}{}
\usepackage[margin=1.0in,bottom=1.5in]{geometry}
\usepackage[square,numbers,sort&compress]{natbib}
\usepackage{listings}
\usepackage{graphicx}
\makeatletter
\def\maxwidth{\ifdim\Gin@nat@width>\linewidth\linewidth\else\Gin@nat@width\fi}
\def\maxheight{\ifdim\Gin@nat@height>\textheight\textheight\else\Gin@nat@height\fi}
\makeatother
\setkeys{Gin}{width=\maxwidth,height=\maxheight,keepaspectratio}
\usepackage{caption}
\usepackage{float}
\usepackage{subfig}
\ifxetex
  \usepackage[setpagesize=false, 
              unicode=false, 
              xetex]{hyperref}
\else
  \usepackage[unicode=true]{hyperref}
\fi
\hypersetup{breaklinks=true,
            bookmarks=true,
            pdfauthor={},
            pdftitle={Faster Uncertainty Quantification for Inverse Problems with Conditional Normalizing Flows},
            colorlinks=true,
            citecolor=black,
            urlcolor=blue,
            linkcolor=black,
            pdfborder={0 0 0}}
\usepackage[preprint]{neurips_2019}
\usepackage{url}
\usepackage{booktabs}
\usepackage{amsfonts}
\usepackage{nicefrac}
\usepackage{microtype}

\newcommand{\KL}{\mathrm{KL}}
\newcommand{\X}{X}
\newcommand{\Y}{Y}
\newcommand{\Zx}{Z_x}
\newcommand{\Zy}{Z_y}
\newcommand{\pXY}{p_{X,Y}}
\newcommand{\pX}{p_{X}}
\newcommand{\pY}{p_{Y}}
\newcommand{\pXcY}{p_{X|Y}}
\newcommand{\pYcX}{p_{Y|X}}
\newcommand{\pZxZy}{p_{Zx,Zy}}
\newcommand{\pZx}{p_{Zx}}
\newcommand{\pZy}{p_{Zy}}
\newcommand{\ppr}{p_{\mathrm{pr}}}

\title{Faster Uncertainty Quantification for Inverse Problems with Conditional
Normalizing Flows}
\author{Ali Siahkoohi\\School of Computational Science and Engineering,\\Georgia
Institute of Technology\\\texttt{alisk@gatech.edu}\\\And
Gabrio Rizzuti\\School of Computational Science and
Engineering,\\Georgia Institute of
Technology\\\texttt{rizzuti.gabrio@gatech.edu}\\\And
Philipp A. Witte\\Formerly Georgia Institute of Technology\\Currently
Microsoft Research\\\texttt{pwitte3@gatech.edu}\\\And
Felix J. Herrmann\\School of Computational Science and
Engineering,\\Georgia Institute of
Technology\\\texttt{felix.herrmann@gatech.edu}}
\date{}

\begin{document}
\maketitle
\begin{abstract}
In inverse problems, we often have access to data consisting of paired
samples $(x,y)\sim \pXY(x,y)$ where $y$ are partial observations of a
physical system, and $x$ represents the unknowns of the problem. Under
these circumstances, we can employ supervised training to learn a
solution $x$ and its uncertainty from the observations $y$. We refer to
this problem as the ``supervised'' case. However, the data $y\sim\pY(y)$
collected at one point could be distributed differently than
observations $y'\sim\pY'(y')$, relevant for a current set of problems.
In the context of Bayesian inference, we propose a two-step scheme,
which makes use of normalizing flows and joint data to train a
conditional generator $q_{\theta}(x|y)$ to approximate the target
posterior density $\pXcY(x|y)$. Additionally, this preliminary phase
provides a density function $q_{\theta}(x|y)$, which can be recast as a
prior for the ``unsupervised'' problem, e.g.~when only the observations
$y'\sim\pY'(y')$, a likelihood model $y'|x$, and a prior on $x'$ are
known. We then train another invertible generator with output density
$q'_{\phi}(x|y')$ specifically for $y'$, allowing us to sample from the
posterior $\pXcY'(x|y')$. We present some synthetic results that
demonstrate considerable training speedup when reusing the pretrained
network $q_{\theta}(x|y')$ as a warm start or preconditioning for
approximating $\pXcY'(x|y')$, instead of learning from scratch. This
training modality can be interpreted as an instance of transfer
learning. This result is particularly relevant for large-scale inverse
problems that employ expensive numerical simulations.
\end{abstract}

\section{Introduction}\label{introduction}

Deep learning techniques have recently benefited inverse problems where
the unknowns defining the state of a physical system and related
observations are jointly available as solution-data paired samples
\citep[see, for example,][]{Adler_2017}. Throughout the text, we will
refer to this problem as the ``supervised'' case. Supervised learning
can be readily applied by training a deep network to map the
observations to the respective solution, often leading to competitive
alternatives to solvers that are purely based on a physical model for
the data likelihood (e.g.~PDEs) and prior (handcrafted) regularization.
Unfortunately, for many inverse problems such as seismic or optoacoustic
imaging, data is scarce due to acquisition costs, processing is
computationally complex because of numerical simulation, and the
physical parameters of interest cannot be directly verified.
Furthermore, as in the seismic case, the vast diversity of geological
scenarios is bound to impact the generalization capacity of the learned
model. For this type of problem, supervised methods have still limited
scope with respect to more traditional ``unsupervised'' approaches,
e.g.~where observations pertaining to a single unknown are available, a
data model and prior are postulated, and generalization errors do not
affect the results. Note that recent work has found an application for
deep networks even in the unsupervised setting as a reparameterization
of the unknowns and an implicit regularizing prior \citep[deep
prior,][]{bora2017compressed, Ulyanov_2020, herrmann2019NIPSliwcuc, siahkoohi2020EAGEdlb, siahkoohi2020SEGwdp, siahkoohi2020SEGhorizonUQ},
by constraining the solution to its range. Unless the network has been
adequately pretrained, however, the deep prior approach does not offer
computational advantages.

In practice, as it is often the case in seismic or medical imaging, some
legacy joint data might be available for supervised learning, while we
might be interested in solving a problem related to some new
observations, which are expected to come from a moderate perturbation of
the legacy (marginal) distribution. In this work, we are interested in
combining the supervised and unsupervised settings, as described above,
by exploiting the supervised result as a way to accelerate the
computation of the solution for the unsupervised problem. Clearly, this
is all the more relevant when we wish to quantify the uncertainty in the
proposed solution.

This paper is based on exploiting conditional normalizing flows
\citep{kruse2019hint, winkler2019learning} as a way to encapsulate the
joint distribution of observations/solution for an inverse problem, and
the posterior distribution of the solutions given data. Recent
advancements have made available invertible flows that allow analytic
computation of such posterior densities. Therefore, we propose a general
two-step scheme which consists of: (\emph{i}) learning a generative
model from many (data, solution) pairs; (\emph{ii}) given some new
observations, we solve for the associated posterior distribution given a
data likelihood model and a prior density (even comprising the one
obtained in step (\emph{i})).

\section{Related work}\label{related-work}

Normalizing flow generative models are the cornerstone of our proposal,
due to their ability to be trained with likelihood-based objectives, and
not being subject to mode collapse. Many invertible layers and
architectures are described in \citet{dinh2014nice},
\citet{dinh2016density}, \citet{kingma2018glow}, and
\citet{kruse2019hint}. A fundamental aspect for their applications to
large-scale imaging problems is constant memory complexity as a function
of the network depth. Examples for seismic imaging can be found in
\citet{peters2020fully} and \citet{rizzuti2020SEGpub}, and for medical
imaging in \citet{putzky2019invert}. In this paper, we will focus on
uncertainty quantification for inverse problems, and we are therefore
interested in the conditional flows described in \citet{kruse2019hint},
as a way to capture posterior probabilities \citep[see
also][]{winkler2019learning}.

Bayesian inference cast as a variational problem is a computationally
attractive alternative to sampling based on Markov chain Monte Carlo
methods (MCMC). With particular relevance for our work,
\citet{marzuk2018} formulates transport-based maps as non-Gaussian
proposal distributions in the context of the Metropolis-Hastings
algorithm. The aim is to accelerate MCMC by adaptively fine-tuning the
proposals to the target density, as samples are iteratively produced by
the chain. The idea of preconditioning MCMC in \citet{marzuk2018}
directly inspires the approach object of this work. Another relevant
work which involve MCMC is \citet{peherstorfer2018transportbased}, where
the transport maps are constructed from a low-fidelity version of the
original problem, thus yielding computational advantages. The supervised
step of our approach can also be replaced, in principle, by a
low-fidelity problem. The method proposed in this paper, however, will
not make use of MCMC.

\section{Method}\label{method}

We start this section by summarizing the uncertainty quantification
method presented in \citet{kruse2019hint}, in the supervised scenario
where paired samples $(x_i,y_i)\sim \pXY(x,y)$ (coming from the joint
unknown/data distribution) are available. We assume that an underlying
physical modeling operator exists, which defines the likelihood model
$\pYcX(y|x)$, $y=F(x)+n$, where $n$ is a random variable representing
noise. The scope is to learn a conditional normalizing flow
\begin{equation}
T:\X\times\Y\to\Zx\times\Zy,
\label{eq:nf}
\end{equation}
 as a way to quantify the uncertainty of the unknown $x$ of an inverse
problem, given data $y$. Here, $(x,y)\in\X\times\Y$, and $\Zx$, $\Zy$
are respective latent spaces. This is achieved by minimizing the
Kullback-Leibler divergence between the push-forward density
$T_{\sharp}\pXY$ and the standard normal distribution
$\pZxZy=\pZxZy(z_x,z_y)=\pZx(z_x)\pZy(z_y)$:
\begin{equation}
\begin{split}
\min_T\KL(T_{\sharp}\pXY||\pZxZy)\hspace{10em}\\=\mathbb{E}_{x,y\sim \pXY(x,y)}\frac{1}{2}||T(x,y)||^2-\log|\det J_T(x,y)|,
\end{split}
\label{eq:klminsup}
\end{equation}
 where $J_T$ is the Jacobian of $T$. When $T$ is a conditional flow,
e.g.~defined by the triangular structure
\begin{equation}
T(x,y)=(T_x(x,y),T_y(y)),
\label{eq:nfcond}
\end{equation}
 conditional sampling given $y$ is tantamount to fixing the data seed
$z_y=T_y(y)$, evaluating $T^{-1}(z_x,z_y)$ for a random Gaussian $z_x$,
and selecting the $x$ component. Moreover, we can analytically evaluate
the approximated posterior density:
\begin{equation}
p_T(x|y)=\pZxZy(T(x,y))|\det J_T(x,y)|\approx \pXcY(x|y).
\label{eq:condpost}
\end{equation}
 We now assume that a map $T$ as in Equation~\eqref{eq:condpost} has
been determined, and we are given new observations $y'\sim \pY'(y')$,
sampled from a marginal $\pY'=\pY'(y')$ closely related to $\pY$. Note
that $y'$ might be obtained with a different forward operator, a
different noise distribution, or an out of prior distribution unknown.
In particular, we assume a different likelihood model
\begin{equation}
p'(y'|x'):\quad y'=F'(x')+n'.
\label{eq:likenew}
\end{equation}
 We are interested in obtaining samples from the posterior
\begin{equation}
\quad \pXcY'(x'|y')=p'(y'|x')\ppr(x'),
\label{eq:unsupcond}
\end{equation}
 with prior $\ppr(x')=\pX(x')$, or even $\ppr(x')=p_T(x'|y')$ as defined
in Equation~\eqref{eq:condpost}, which corresponds to reusing the
supervised posterior as the new prior. Similarly to the previous step,
we can setup a variational problem
\begin{equation}
\begin{split}
\min_{S}\KL(S_{\sharp}\pZx||\pXcY'(\cdot|y'))\hspace{10em}\\=\mathbb{E}_{z_x\sim\pZx(z_x)}-\log\pXcY'(S(z_x)|y')-\log|\det J_S(z_x)|,
\end{split}
\label{eq:klminunsup}
\end{equation}
 where we minimize over the set of invertible maps
\begin{equation}
S:\Zx\to\X.
\label{eq:nfcondunsup}
\end{equation}
 After training, samples from $\pXcY'(x|y')$ are obtained by evaluating
$S(z_x)$ for $z_x\sim\pZx(z_x)$.

For the problem in Equation~\eqref{eq:klminunsup}, we can initialize the
network $S=S_0$ randomly. However, if we expect the supervised
problem~\eqref{eq:klminsup} and the unsupervised
counterpart~\eqref{eq:klminunsup} to be related, we can reuse the
supervised result $T$ as a warm start for $S$, e.g.
\begin{equation}
S_0(z_x)=\pi_{\X}\circ T^{-1}(z_x,z_y),
\label{eq:warmstart}
\end{equation}
 where $\pi_{\X}(x,y)=x$ is the projection on $\X$. By doing so, we can
be interpreted the problem in Equation~\eqref{eq:warmstart} as an
instance of transfer learning
\citep{yosinski2014transferable, siahkoohi2019transfer}. Alternatively,
by analogy with the technique of preconditioning for linear systems, we
can introduce the change of variable
\begin{equation}
S(z_x)=\bar{S}\circ\pi_{\X}\circ T^{-1}(z_x,z_y),\quad\bar{S}:\X\to\X,
\label{eq:cov}
\end{equation}
 and solve for $\bar{S}$ instead of $S$.

\section{Numerical experiments}\label{numerical-experiments}

In this section, we present some synthetic examples aimed at verifying
the speed up anticipated from the two step preconditioning. The first
example is a low-dimensional problem where the posterior density can be
calculated analytically, with which we can ascertain our solution. The
second example constitutes a preliminary assessment for the type of
inverse problem applications we are mostly interested in, e.g.~seismic
or optoacoustic imaging.

\subsection{Gaussian likelihood and
prior}\label{gaussian-likelihood-and-prior}

Here, we consider unknowns $x\in\mathbb{R}^{N_x}$ with $N_x=12$. The
prior density $\pX=\pX(x)$ is a normal distribution
$\pX=\mathcal{N}(\mu_x,\Sigma_x)$ with $\mu_x=1$ and
$\Sigma_x=\mathrm{diag}(1,2,\ldots,12)$. Observations are
$y\in\mathbb{R}^{N_y}$ with $N_y=6$, and we consider the following
likelihood model $\pYcX=\pYcX(y|x)$:
\begin{equation}
y=A\,x+\varepsilon,\quad\varepsilon\sim \mathcal{N}(\mu_{\varepsilon},\Sigma_{\varepsilon}).
\label{eq:ex1like}
\end{equation}
 Mean and covariance are chosen to be $\mu_{\varepsilon}=0$ and
$\Sigma_{\varepsilon}=0.1\,I$ ($I$ being the identity matrix). The
forward operator $A\in\mathrm{Mat}_{N_y,N_x}(\mathbb{R})$ is a
realization of a random matrix variable with independent entries
distributed accordingly to $a_{ij}\sim\mathcal{N}(0,1/N_x)$. We trained
a conditional invertible network to jointly sample from
$(x,y)\sim\pXY(x,y)$.

Let us assume now that new observations $y'$ have been collected. We
generated those observations according to
\begin{equation}
y'=A\,x'+\varepsilon',\quad\varepsilon'\sim \mathcal{N}(\mu_{\varepsilon'},\Sigma_{\varepsilon'})
\label{eq:ex1likeNew}
\end{equation}
 with $x'\sim\mathcal{N}(\mu_{x'},\Sigma_{x'})$, $\mu_{x'}=3\mu_x$,
$\Sigma_{x'}=1.96\,\Sigma_x^{0.3}$, and same noise distribution as
before $\mu_{\varepsilon'}=\mu_{\varepsilon}$,
$\Sigma_{\varepsilon'}=\Sigma_{\varepsilon}$. The likelihood model for
$y'$, in conjunction with the same prior $\pX=\pX(x')$ as in the
supervised case, defines the unsupervised problem.

\begin{figure}
\centering
\subfloat[\label{figCompGauss-a}]{\includegraphics[width=0.700\hsize]{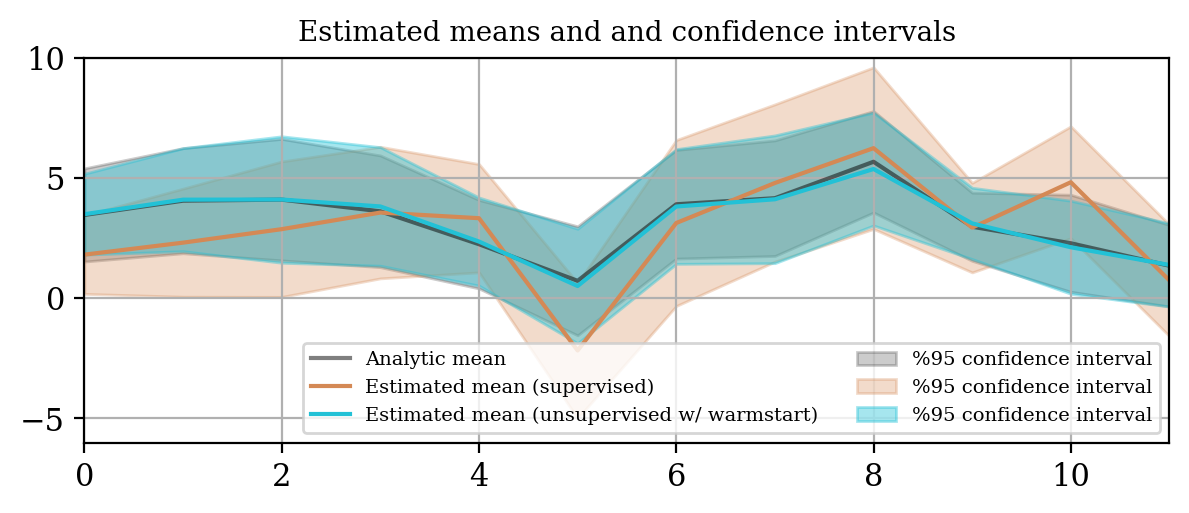}}
\\
\subfloat[\label{figCompGauss-b}]{\includegraphics[width=0.333\hsize]{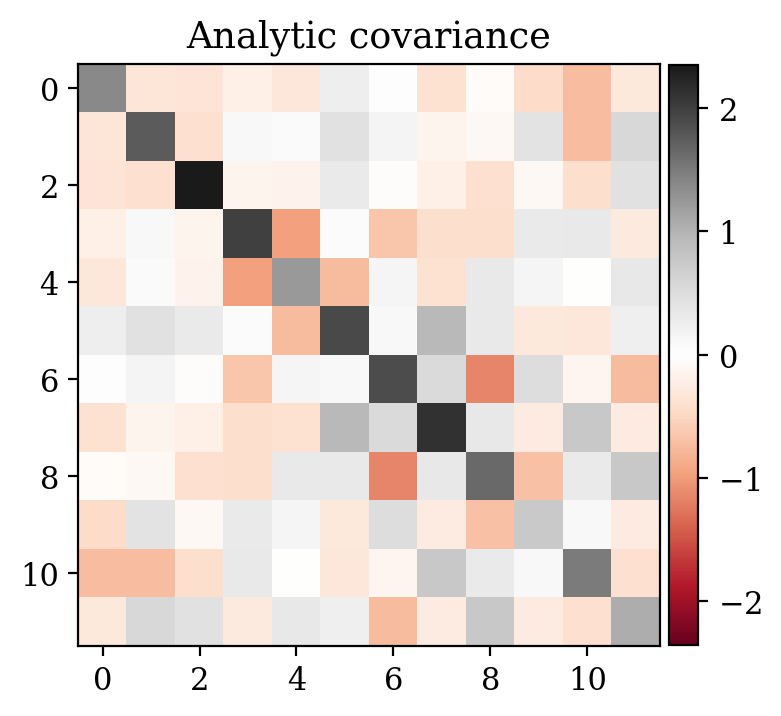}}
\subfloat[\label{figCompGauss-c}]{\includegraphics[width=0.333\hsize]{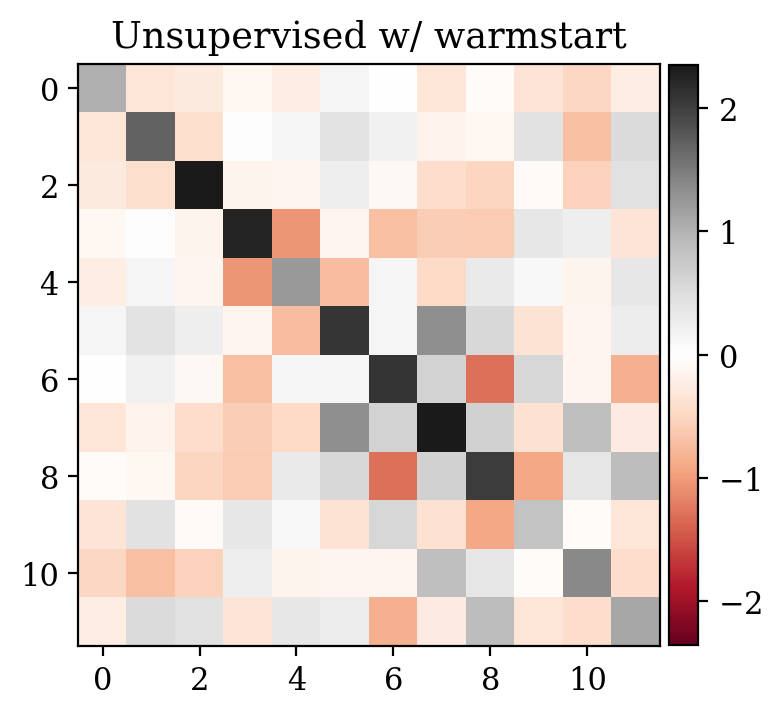}}
\subfloat[\label{figCompGauss-d}]{\includegraphics[width=0.333\hsize]{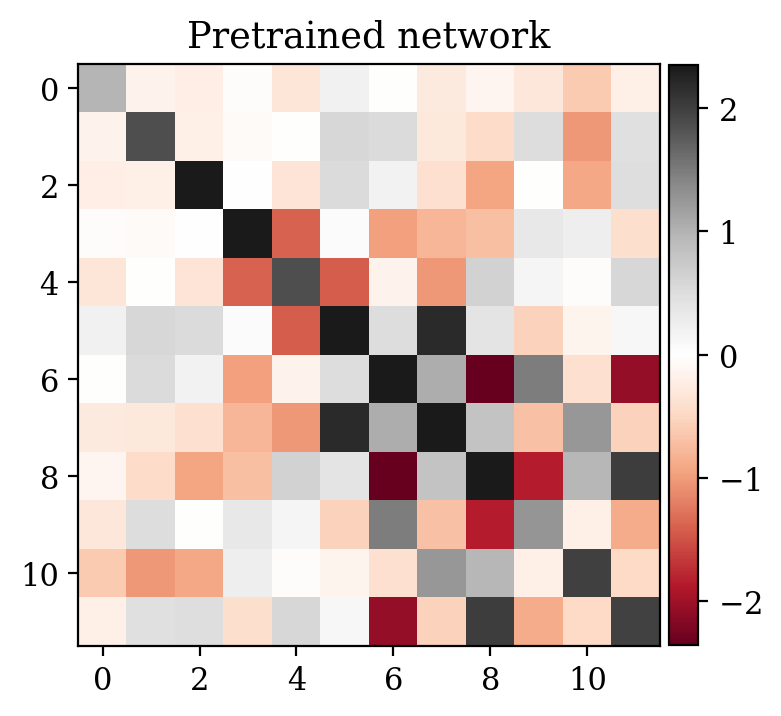}}
\caption{Comparison of the posterior mean (a) and covariance (b--d)
obtained from supervised and unsupervised training for the Gaussian
problem. Note that the results are supposed to differ, due to different
prior and observation models. The analytic mean and covariance here
refers to the unsupervised problem.}\label{figCompGauss}
\end{figure}

The uncertainty quantification results for the
supervised~\eqref{eq:ex1like} and unsupervised
problem~\eqref{eq:ex1likeNew} are compared in Figure~\ref{figCompGauss}.

\begin{figure}
\centering
\includegraphics[width=0.700\hsize]{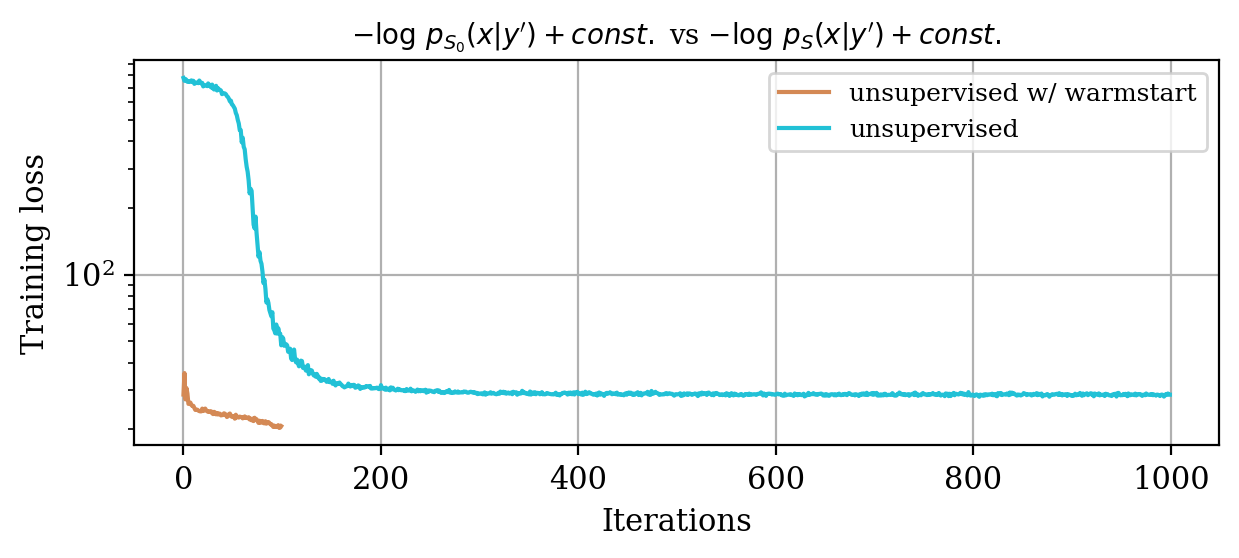}
\caption{Loss decay for the unsupervised Gaussian problem as a function
of iterations. The comparison highlights the efficiency of the warm
start strategy compared to training from scratch.}\label{figLossGauss}
\end{figure}

We study the convergence history for unsupervised training with and
without warm start, as described in Equation~\eqref{eq:warmstart}. The
plot in Figure~\ref{figLossGauss} makes clear the computational
superiority of the warm start approach.

\subsection{Seismic images}\label{seismic-images}

Now we consider the denoising problem for 2D ``seismic'' images $x$,
which are selected from the processed 3D seismic survey reported in
\citet{Veritas2005} and \citet{WesternGeco2012}. The dataset has been
obtained by selecting 2D patches from the original 3D volume, which are
then subsampled in order to obtain $64\times 64$ pixel images. The
dataset is normalized.

\begin{figure}
\centering
\captionsetup[subfigure]{labelformat=empty}
\subfloat[]{\includegraphics[width=0.333\hsize]{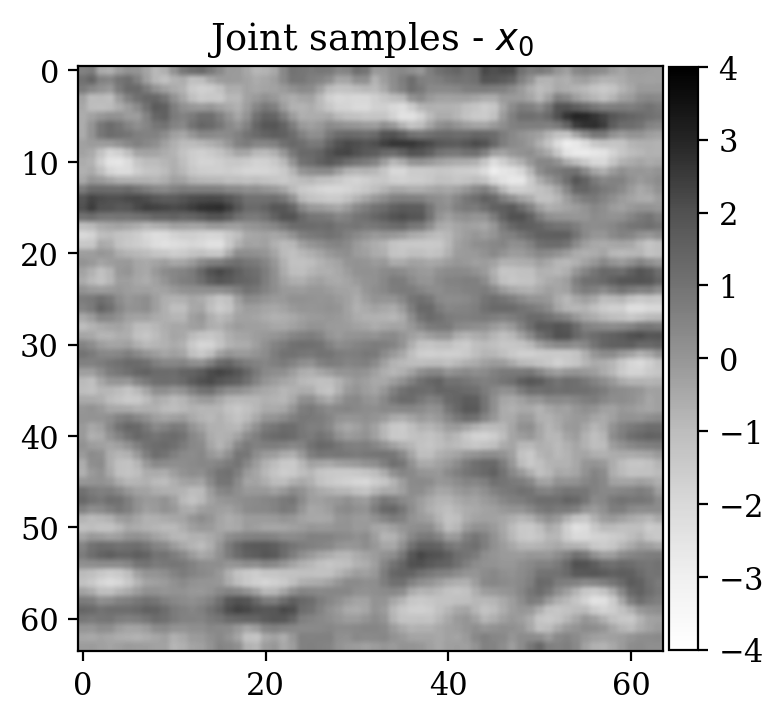}}
\subfloat[]{\includegraphics[width=0.333\hsize]{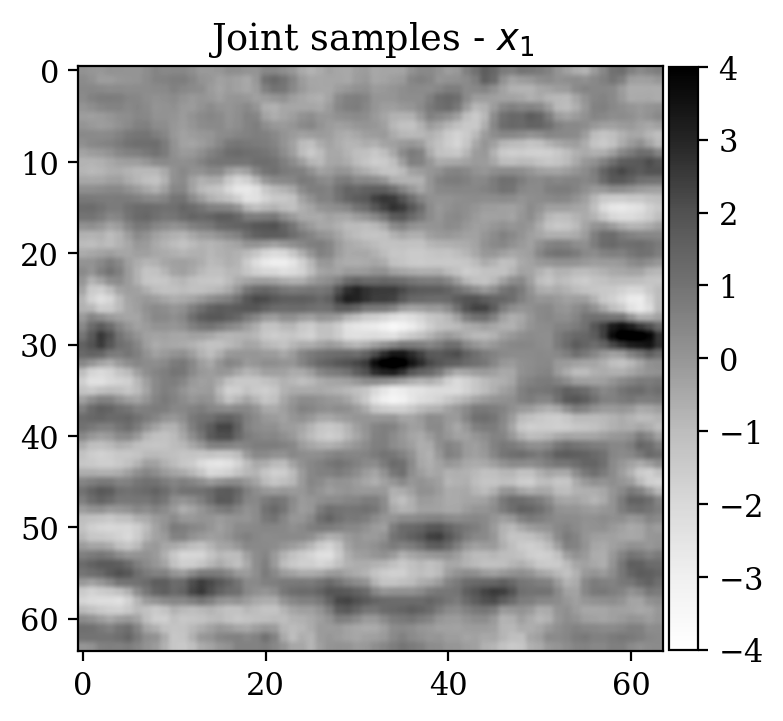}}
\subfloat[]{\includegraphics[width=0.333\hsize]{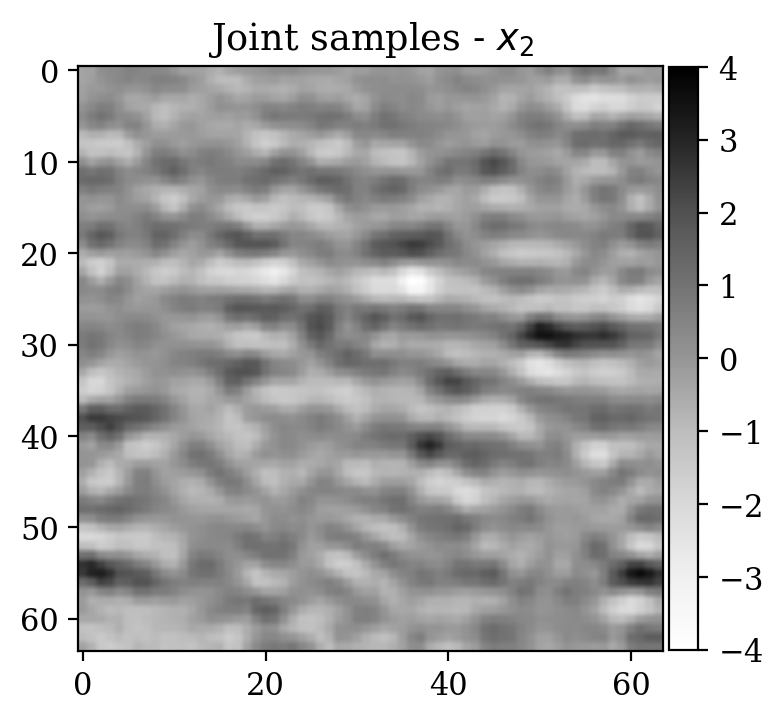}}
\\
\subfloat[]{\includegraphics[width=0.333\hsize]{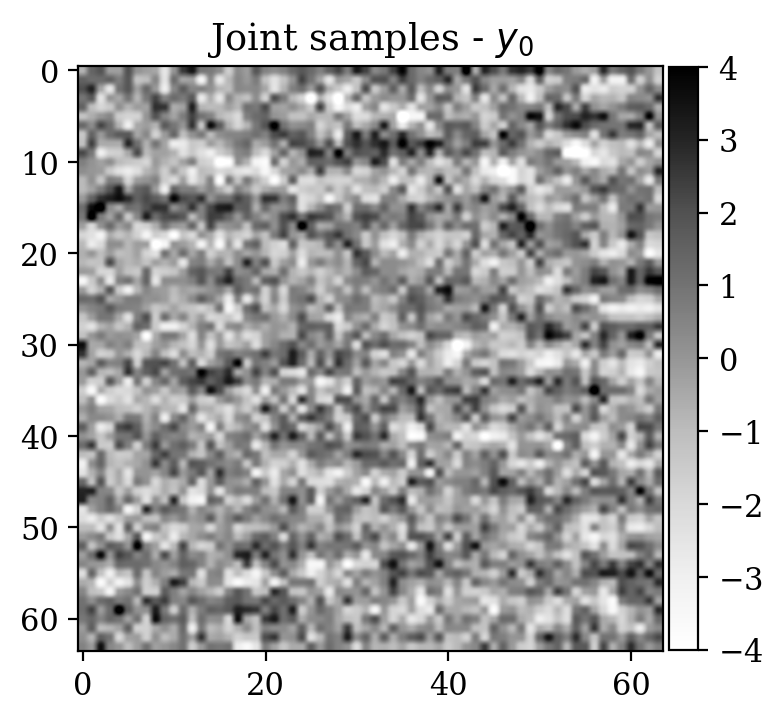}}
\subfloat[]{\includegraphics[width=0.333\hsize]{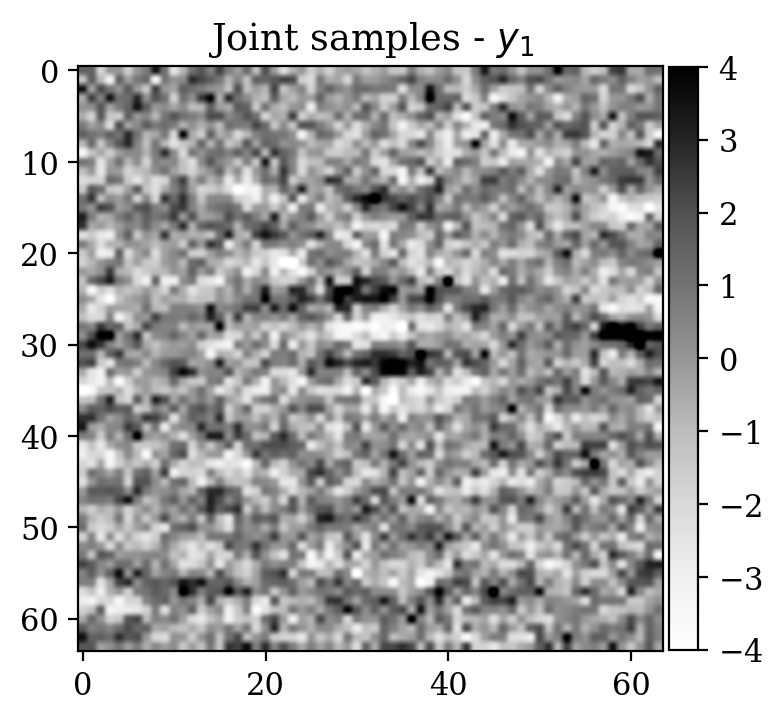}}
\subfloat[]{\includegraphics[width=0.333\hsize]{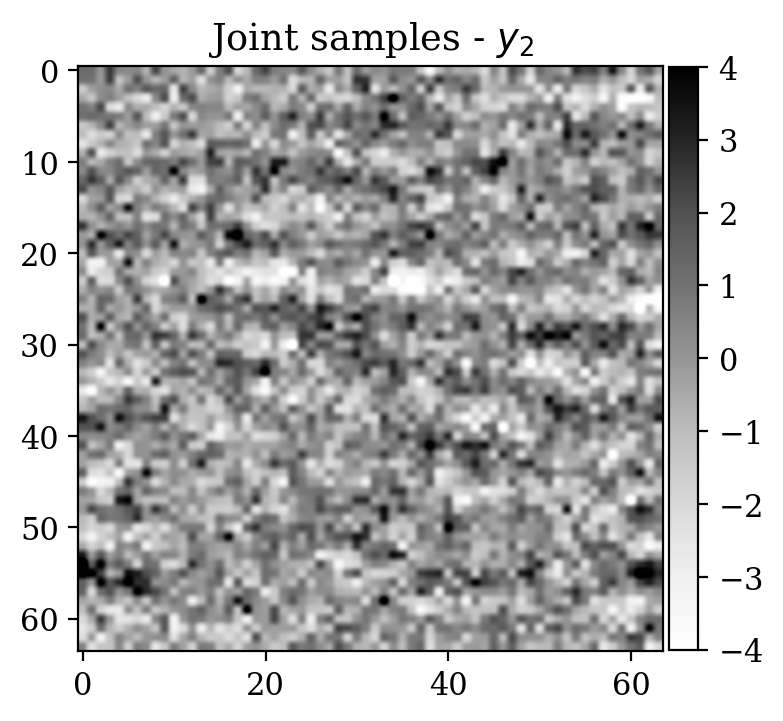}}
\caption{Unknown and observation pairs $(x,y)$ for the supervised
seismic image problem (first row: $x$, second row:
$y$).}\label{figSeisImage}
\end{figure}

Observations $y$ are obtained simply by adding noise
\begin{equation}
y=x+\varepsilon,\quad\varepsilon\sim \mathcal{N}(\mu_{\varepsilon},\Sigma_{\varepsilon})
\label{eq:ex2like}
\end{equation}
 with $\mu_{\varepsilon}=0$ and $\Sigma_{\varepsilon}=1.2\,I$. Examples
of $(x,y)$ pairs are collected in Figure~\ref{figSeisImage}. As in the
previous examples, we consider a preliminary stage for supervised
training via conditional normalizing flows.

We now turn to the unsupervised problem defined by the observation
likelihood
\begin{equation}
y'=A\,x'+\varepsilon',\quad\varepsilon'\sim \mathcal{N}(\mu_{\varepsilon'},\Sigma_{\varepsilon'})
\label{eq:ex2like2}
\end{equation}
 with $\mu_{\varepsilon'}=0$ and $\Sigma_{\varepsilon'}=0.2\,I$. Note
that a forward operator $A$ has been introduced, contrary to
Equation~\eqref{eq:ex2like}. Here, $A$ is equal to $B^T B$, where $B$ is
a compressing sensing matrix with $30 \%$ subsampling rate. The ground
truth $x'$ for observations $y'$ has been selected from a test set not
contemplated during the supervised training phase. As a prior for $x'$,
we select the posterior distribution given $y'$ which has been
pretrained with supervision in the previous step (see
Equation~\eqref{eq:condpost}).

\begin{figure}
\centering
\subfloat[\label{figLossSeis-a}]{\includegraphics[width=0.700\hsize]{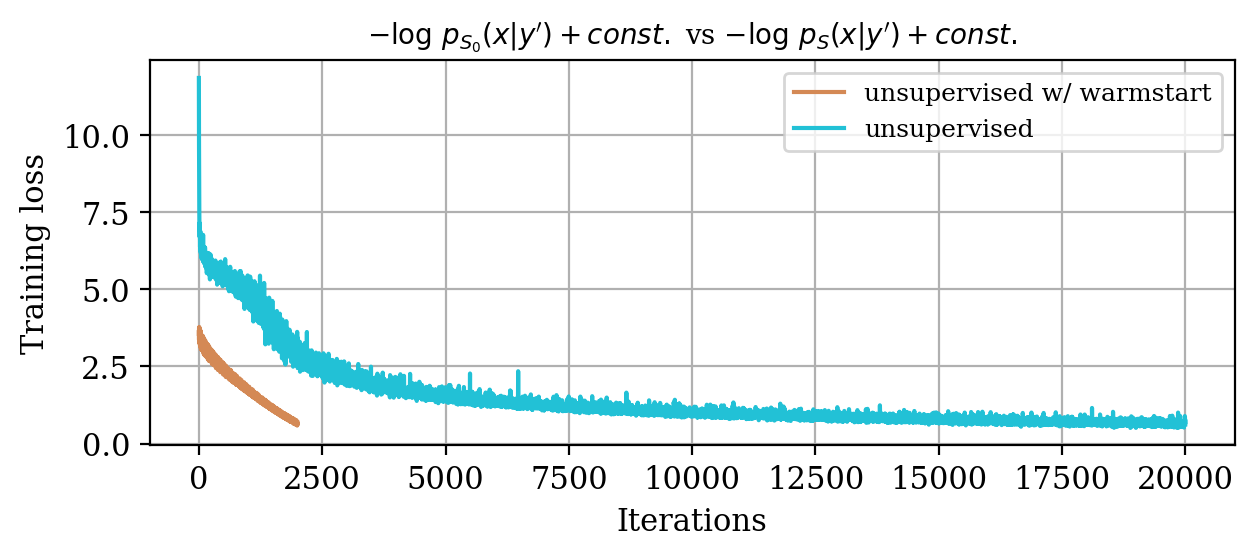}}
\\
\subfloat[\label{figLossSeis-b}]{\includegraphics[width=0.700\hsize]{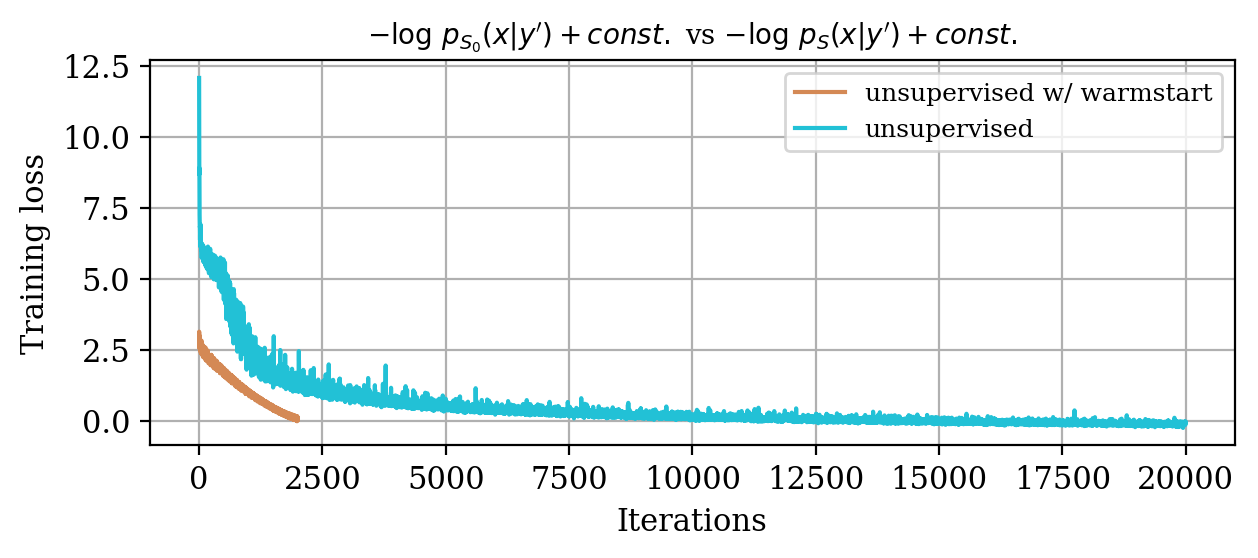}}
\caption{Loss decays for two different instances of the unsupervised
problems related to seismic images. As in the previous example, training
with a warm start strategy evidently requires less iterations than
training from scratch in order to reach the same loss
value.}\label{figLossSeis}
\end{figure}

Again, comparing the loss decay during training for two different
instances of the unsupervised problem in Figure~\ref{figLossSeis}, makes
clear that considerable speed up is obtained with the warm start
strategy relatively to training a randomly initialized invertible
network.

\begin{figure}
\centering
\subfloat[\label{figCompSeis-a}]{\includegraphics[width=0.380\hsize]{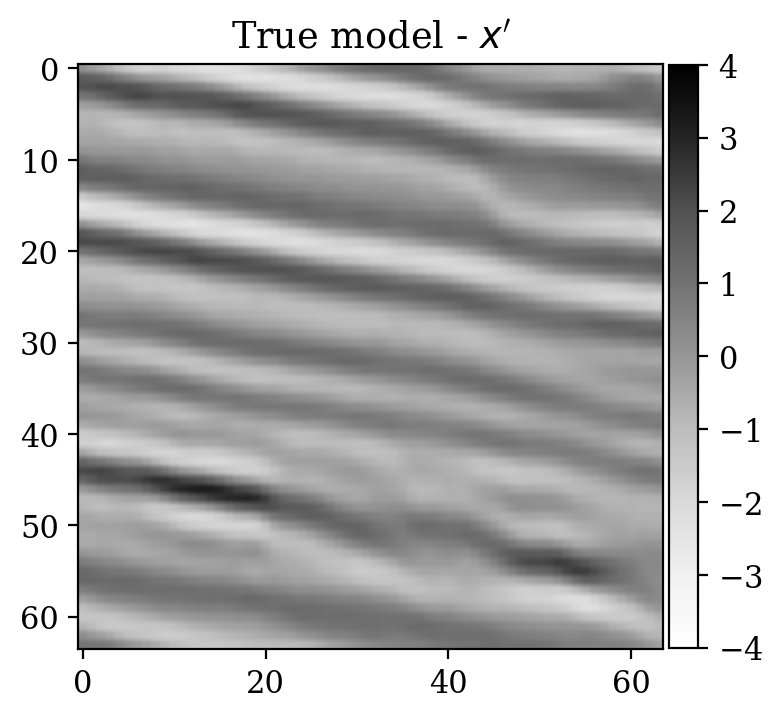}}
\subfloat[\label{figCompSeis-b}]{\includegraphics[width=0.380\hsize]{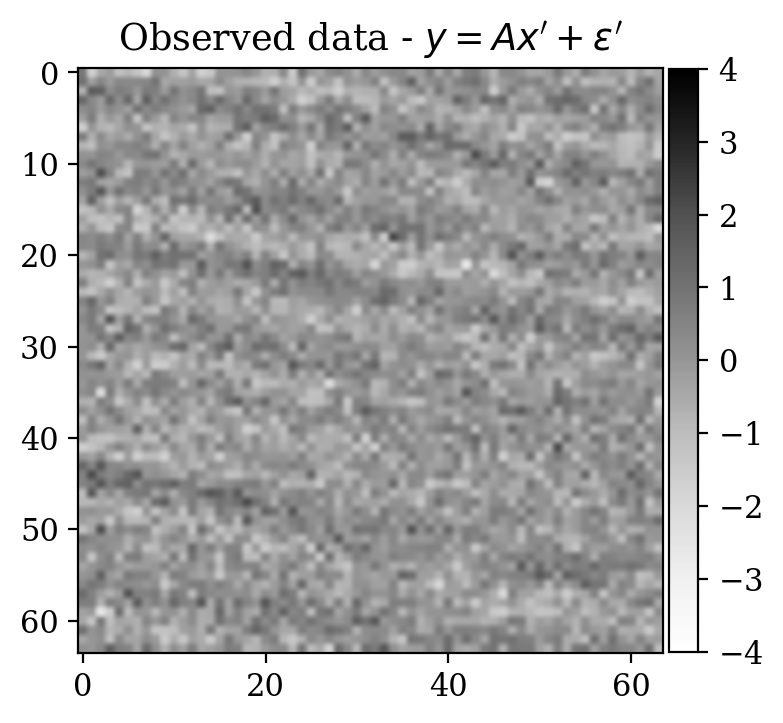}}
\\
\subfloat[\label{figCompSeis-c}]{\includegraphics[width=0.380\hsize]{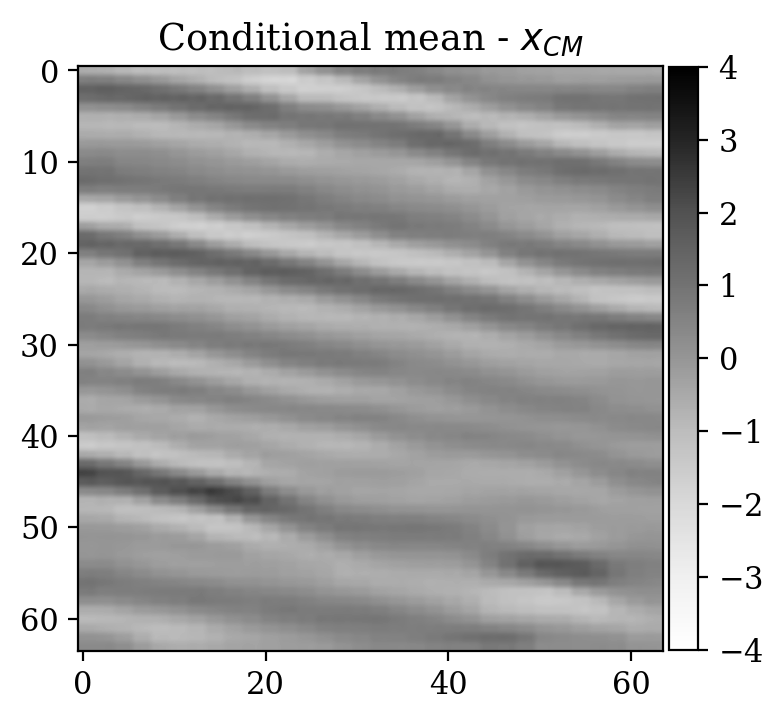}}
\subfloat[\label{figCompSeis-d}]{\includegraphics[width=0.380\hsize]{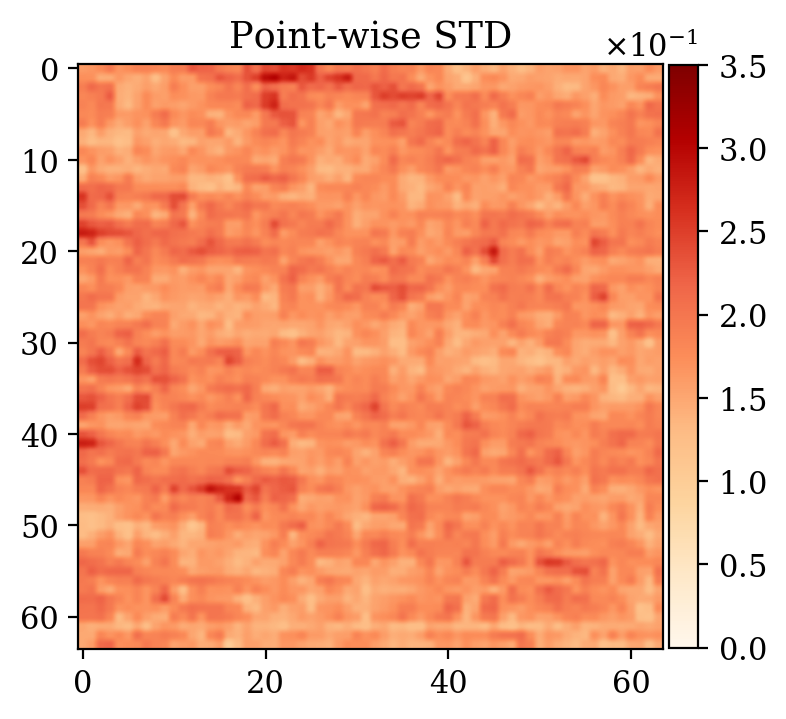}}
\\
\subfloat[\label{figCompSeis-e}]{\includegraphics[width=0.380\hsize]{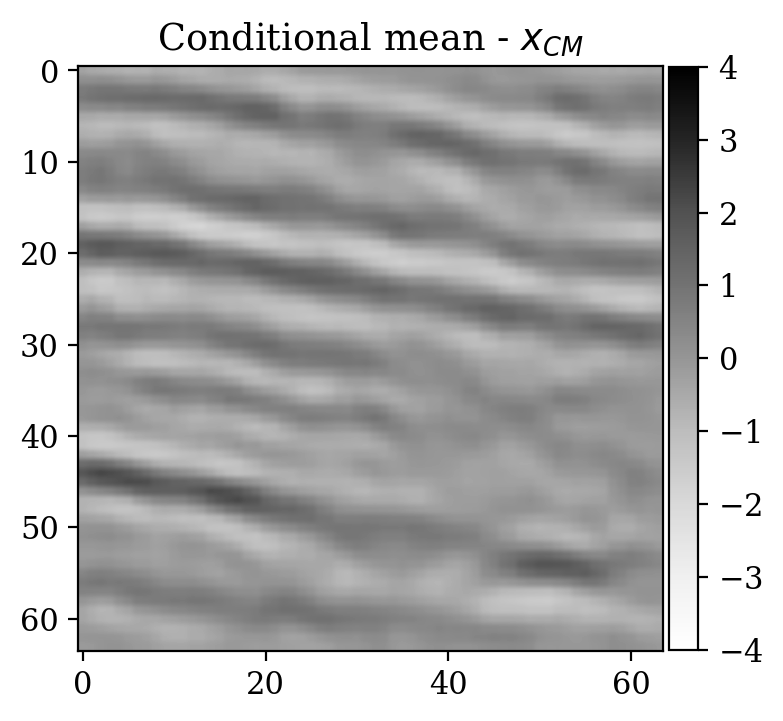}}
\subfloat[\label{figCompSeis-f}]{\includegraphics[width=0.380\hsize]{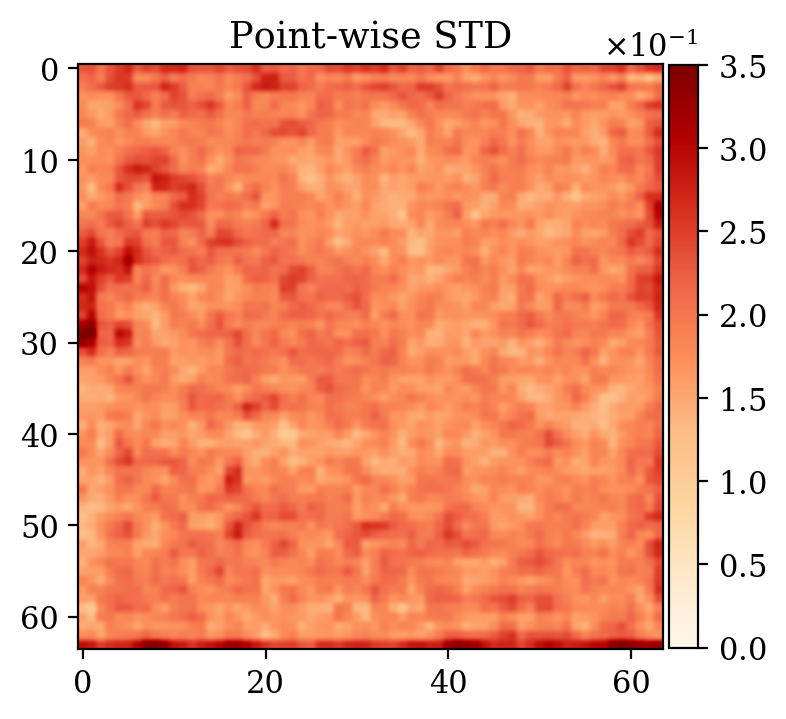}}
\caption{Comparison of the posterior distribution obtained from training
a network with warm start and from scratch for the unsupervised seismic
image problem. Figures (a) and (b) depict the ground truth and related
observations. Figures (c) and (e) represent the respective conditional
means, while (d) and (f) refer to the pointwise standard deviation. Note
how the result in (c) provides a better estimation of the ground truth
(a) compared to (e).}\label{figCompSeis2}
\end{figure}

Moreover, despite the relatively high number of iterations ran during
training ($\sim20000$), the network initialized from scratch does not
produce a reasonable result comparable to the ground truth. This can be
seen by comparing the ground truth in Figure~\ref{figCompSeis-a} and the
conditional mean in Figure~\ref{figCompSeis-e}, relative to the
posterior distribution obtained from training a network from scratch.
The comparison with the ground truth is much more favorable with the
conditional mean obtained from the warm start training, in
Figure~\ref{figCompSeis-c}. Pointwise standard deviations for these
different training modalities can also be inspected in
Figures~\ref{figCompSeis-d} (warm start) and~\ref{figCompSeis-f}
(without warm start). The discussed results above are related to the
loss function depicted in Figure~\ref{figLossSeis-a}. Same results for a
different realization of the unsupervised problem with loss function
shown in Figure~\ref{figLossSeis-b} can be seen in
Figure~\ref{figCompSeis2}.

\begin{figure}
\centering
\subfloat[\label{figCompSeis-a}]{\includegraphics[width=0.380\hsize]{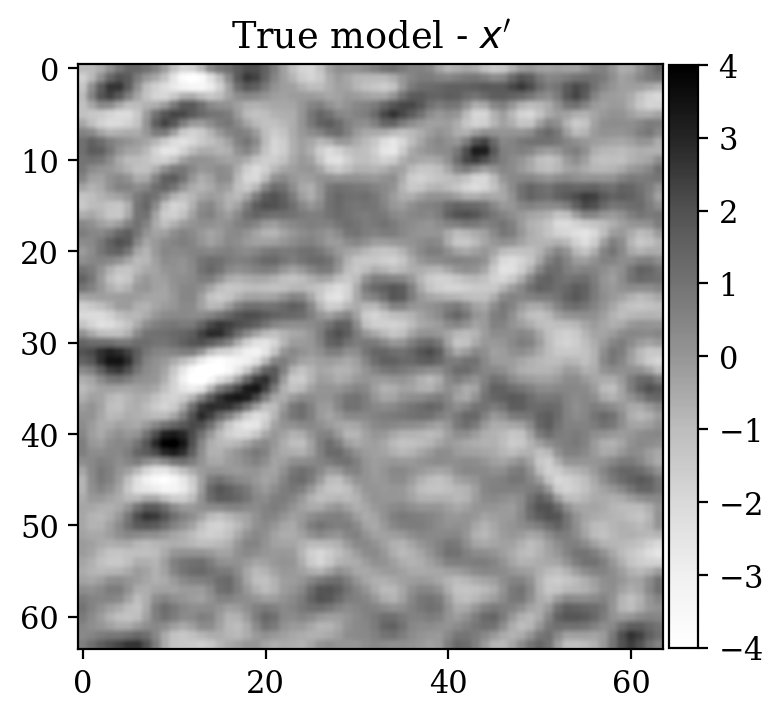}}
\subfloat[\label{figCompSeis-b}]{\includegraphics[width=0.380\hsize]{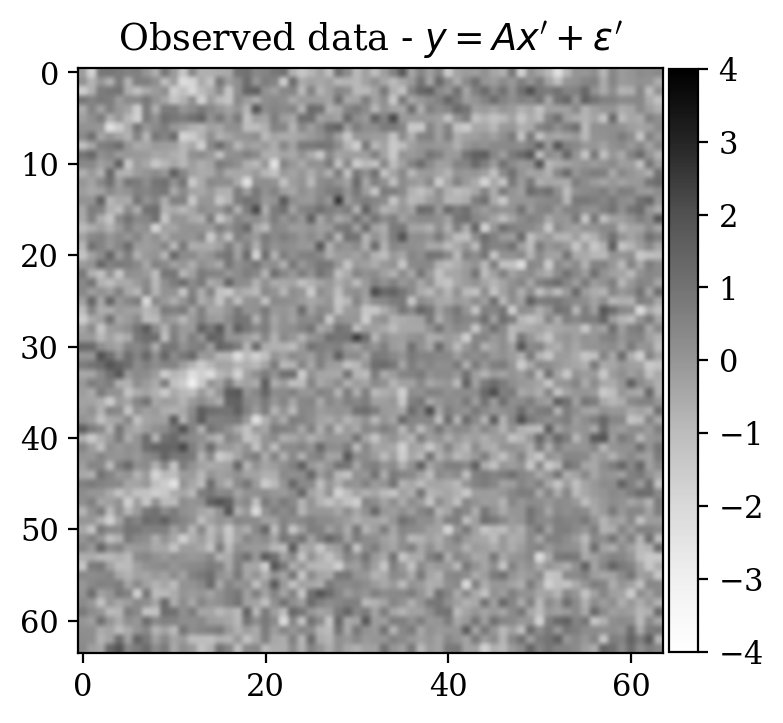}}
\\
\subfloat[\label{figCompSeis-c}]{\includegraphics[width=0.380\hsize]{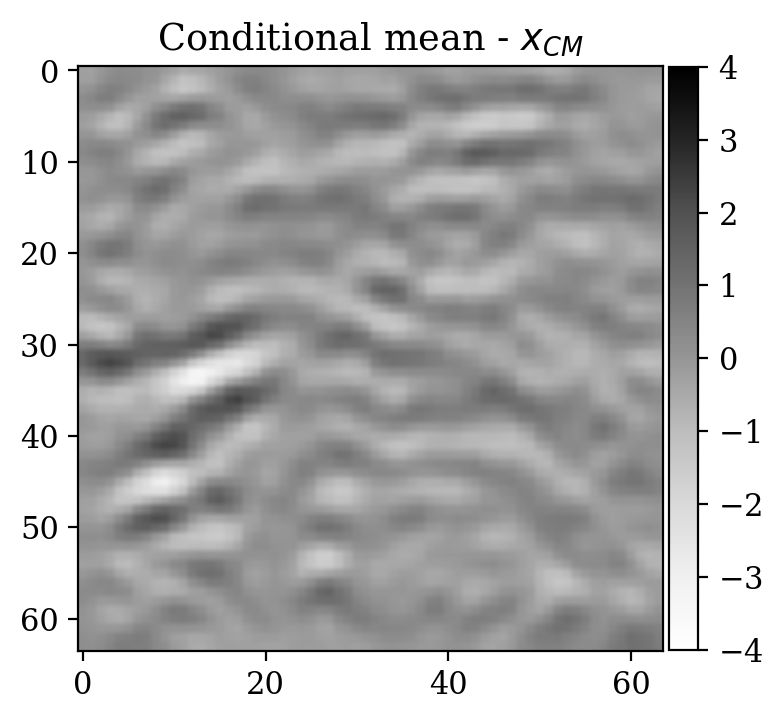}}
\subfloat[\label{figCompSeis-d}]{\includegraphics[width=0.380\hsize]{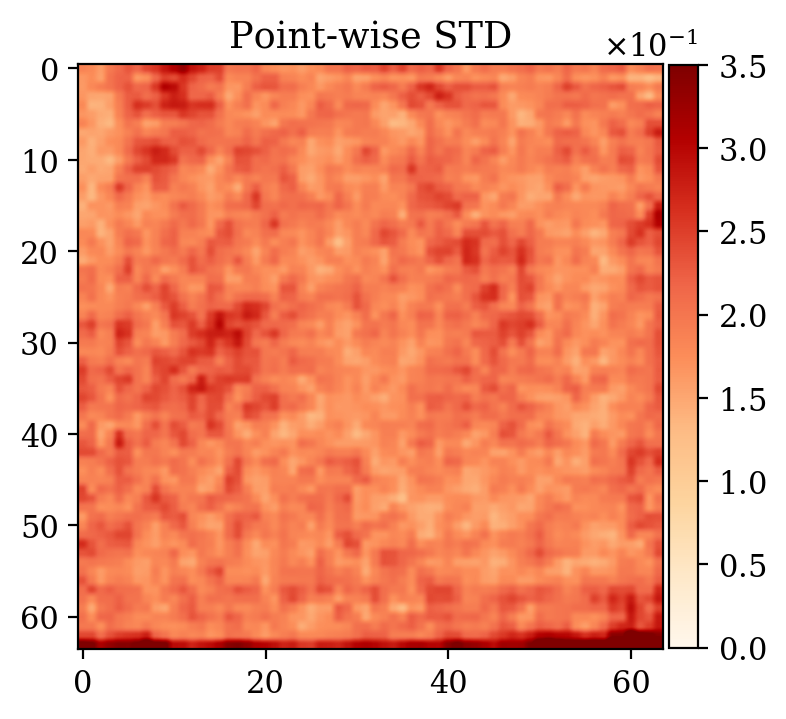}}
\\
\subfloat[\label{figCompSeis-e}]{\includegraphics[width=0.380\hsize]{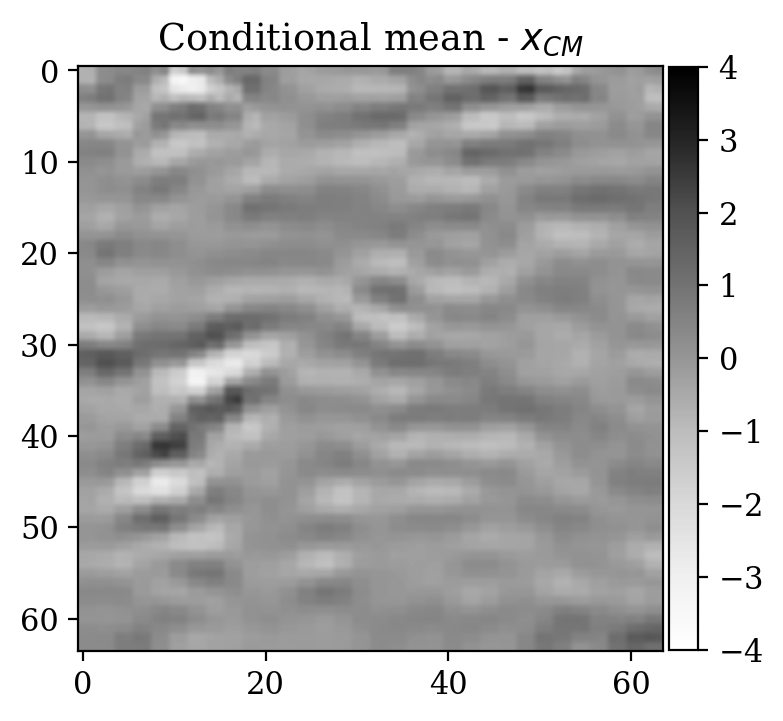}}
\subfloat[\label{figCompSeis-f}]{\includegraphics[width=0.380\hsize]{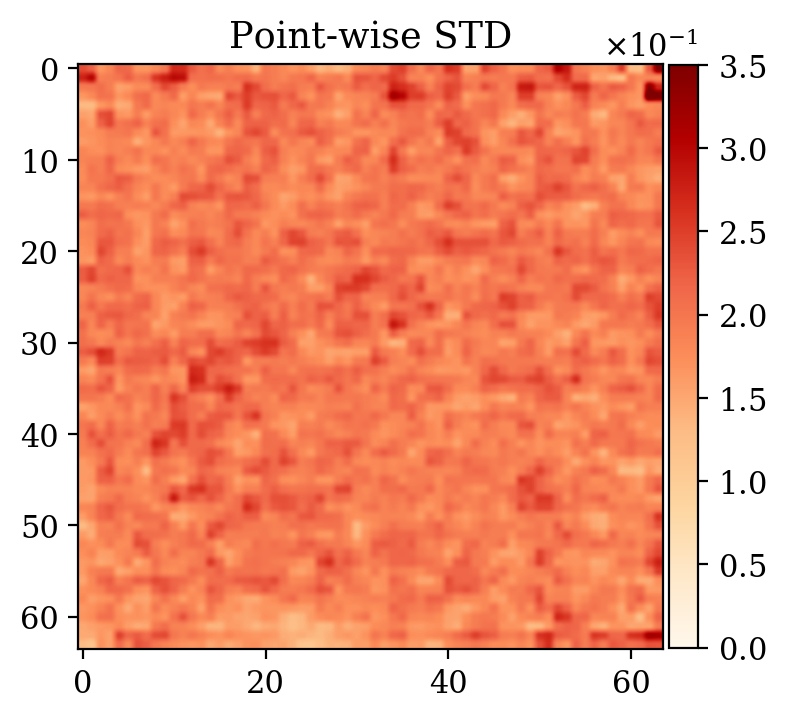}}
\caption{Comparison of the posterior distribution obtained from training
a network with warm start and from scratch for the unsupervised seismic
image problem. Figures (a) and (b) depict the ground truth and related
observations. Figures (c) and (e) represent the respective conditional
means, while (d) and (f) refer to the pointwise standard deviation. Note
how the result in (c) provides a better estimation of the ground truth
(a) compared to (e).}\label{figCompSeis}
\end{figure}

\section{Conclusions}\label{conclusions}

We presented a preconditioning scheme for uncertainty quantification,
particularly aimed at inverse problems characterized by computationally
expensive numerical simulations based on PDEs (including, for example,
seismic or optoacoustic imaging). We consider the problem where legacy
supervised data is available, and we want to solve for a new inverse
problem given some out-of-distribution observations. The scheme takes
advantage of a preliminary step where the joint distribution of solution
and related observations is learned via supervised learning. This joint
distribution is then employed as a way to precondition the unsupervised
inverse problem. In the supervised and unsupervised case, we make use of
conditional normalizing flows to ease computational complexity
(fundamental for large 3D applications), and to be able to encode
analytically the approximated posterior density. In this way, the
posterior density obtained from the supervised problem can be reused as
a new prior for the unsupervised problem.

The synthetic experiments confirm that the preconditioning scheme
accelerates unsupervised training considerably. The examples here
considered are encouraging for seismic or optoacoustic imaging
applications, but additional challenges are expected for large scales
due to the high dimensionality of the solution and observation space,
and expensive wave equation solvers.

\bibliography{biblio_rizzuti2020ICMLws}

\end{document}